# ChatGPT as the Marketplace of Ideas: Should Truth-Seeking Be the Goal of AI Content Governance?

Jiawei Zhang[*]

*As one of the most enduring metaphors within legal discourse, the marketplace of ideas has wielded considerable influence over the jurisprudential landscape for decades. A century after the inception of this theory, ChatGPT emerged as a revolutionary technological advancement in the twenty-first century. This research finds that ChatGPT effectively manifests the marketplace metaphor—it not only instantiates the promises envisaged by generations of legal scholars but also lays bare the perils discerned through sustained academic critique. Specifically, the workings of ChatGPT and the marketplace of ideas theory exhibit at least four common features: arena, means, objectives, and flaws. These shared attributes are sufficient to render ChatGPT historically the most qualified engine for actualizing the marketplace of ideas theory.*

*The comparison of the marketplace theory and ChatGPT merely marks a starting point. A more meaningful undertaking entails reevaluating and reframing both internal and external AI policies by referring to the accumulated experience, insights, and suggestions researchers have raised to fix the marketplace theory. Here, a pivotal issue is: should truth-seeking be set as the goal of AI content governance? Given the unattainability of the absolute truth-seeking goal, I argue against adopting zero-risk policies. Instead, a more judicious approach would be to embrace a knowledge-based alternative wherein large language models (LLMs) are trained to generate*

[*] Ph.D. Candidate and Research Assistant at the Technical University of Munich, Guest Researcher at Max Planck Institute for Innovation and Competition; M.Phil. (Oxon), LL.M. (U.C. Berkeley). Many thanks to Urs Gasser, Yuan Hao, Margot Kaminski, Xiangyu Ma, Helga Nowotny, Boris Paal, Pamela Samuelson, Chongyao Wang, Guangrun Wang, Klaus Wiedemann, and Stefan Wurster for their time in reviewing drafts of this work and valuable feedback provided at different stages of this Essay; to Nina Grgić-Hlača, Anselm Küsters, Gabriel Lima, Julius Schumann, Frederike Zufall, and other organizers and participants at the 3rd Max Planck Law | Tech | Society Graduate Student Symposium for their organization, participation, and great questions; and to Makena A. Kauhane, Leo Rassieur, Marissa Cheng Uri, and other editors of the *Stanford Law & Policy Review* for their diligent and excellent editorial work on this Essay. All errors are my own. I welcome any comments on this Essay: jiawei.zhang@tum.de.

*competing and divergent viewpoints based on sufficient justifications. This research also argues that so-called AI content risks are not created by AI companies but are inherent in the entire information ecosystem. Thus, the burden of managing these risks should be distributed among different social actors, rather than being solely shouldered by chatbot companies.*

*Table of Contents*

> [T]he ultimate good desired is better reached by free trade in ideas — that the best test of *truth* is the power of the thought to get itself accepted in the competition of the market.
>
> — Oliver Wendell Holmes (1919)[1]

> I'm going to start something which I call TruthGPT or a maximum *truth-seeking AI* that tries to understand the nature of the universe . . . [a]nd I think this might be the best path to safety in the sense that an AI that cares about understanding

---

[1] Abrams v. United States, 250 U.S. 616, 630 (1919) (Holmes, J., dissenting) (emphasis added).

> the universe is unlikely to annihilate humans because we are an interesting part of the universe.
>
> — Elon Musk (2023)[2]

## I. INTRODUCTION

In 1919, Justice Oliver Wendell Holmes Jr.'s dissent in *Abrams v. United States* introduced the notion of the "marketplace of ideas," albeit without using this terminology directly.[3] The core spirit of this metaphor is that the public interest can be best served when the invisible hand of the information market can filter out misinformation and find the truth that ultimately prevails in that market. This rationale has been widely invoked by courts to support free speech doctrines and discourage unwarranted governmental interferences in the information marketplace.[4] Over the ensuing century, this theory has also been fervently discussed in academia.[5] Scholars have now reached a consensus that the marketplace of ideas proves

---

[2] Emma Roth, *Elon Musk Claims to Be Working on 'TruthGPT' — A 'Maximum Truth-Seeking AI'*, VERGE (Apr. 18, 2023) (emphasis added), https://perma.cc/Z2VH-VZYA.

[3] *See Abrams*, 250 U.S. at 630 (Holmes, J., dissenting).

[4] *See* Rodney A. Smolla, *The Meaning of the "Marketplace of Ideas" in First Amendment Law*, 24 COMMC'N L. & POL'Y 437, 439-41 (2019).

[5] *See generally, e.g.*, R. H. Coase, *Advertising and Free Speech*, 6 J. LEGAL STUD. 1 (1977); Stanley Ingber, *The Marketplace of Ideas: A Legitimizing Myth*, 1984 DUKE L.J. 1 (1984); Alvin I. Goldman & James C. Cox, *Speech, Truth, and the Free Market for Ideas*, 2 LEGAL THEORY 1 (1996); Paul H. Brietzke, *How and Why the Marketplace of Ideas Fails*, 31 VAL. U. L. REV. 951 (1996); Vincent Blasi, *Holmes and the Marketplace of Ideas*, 2004 SUP. CT. REV. 1; Derek E. Bambauer, *Shopping Badly: Cognitive Biases, Communications, and the Fallacy of the Marketplace of Ideas*, 77 U. COLO. L. REV. 649 (2006); Joseph Blocher, *Institutions in the Marketplace of Ideas*, 57 DUKE L.J. 821 (2007); Daniel E. Ho & Frederick Schauer, *Testing the Marketplace of Ideas*, 90 N.Y.U. L. REV. 1160 (2015); Dawn Carla Nunziato, *The Marketplace of Ideas Online*, 94 NOTRE DAME L. REV. 1519 (2018); G. Michael Parsons, *Fighting for Attention: Democracy, Free Speech, and the Marketplace of Ideas*, 104 MINN. L. REV. 2157 (2019).

difficult to implement in practice and the ideal of truth-seeking is unattainable in the real world.[6]

Today, the advent of generative AI systems has physicalized the marketplace of ideas. In late 2022, OpenAI released ChatGPT, which soon swept across the world.[7] AI chatbots like ChatGPT[8] learn to generate human-like language to respond to users' prompts by employing "next-token prediction."[9] This fundamental concept used in large language models (LLMs) involves predicting the most likely next word in a sequence of words based on the previous text.[10] Such deep learning models, particularly those based on "transformer architecture," excel at identifying intricate patterns within data.[11] Their ability to perform next-token prediction significantly improves the precision and efficiency of continuous text-generation processes.[12] Beyond statistical techniques, GPT models are evolving into autonomous AI systems. For instance, GPT-4 can perform web browsing and utilize various software tools, including calling on other artificial intelligence models.[13] To date, ChatGPT has brought about pressing concerns over AI-generated content risks, including harmful content, discrimination and bias, and misinformation.[14] These problematic outputs have triggered a potpourri of regulatory proposals worldwide.[15]

---

[6] *See, e.g.*, Ingber, *supra* note 5, at 48 (arguing that "[t]he marketplace of ideas is more myth than reality"); Parsons, *supra* note 5, at 2159 (asserting that "the marketplace of ideas rests upon little more than slogans and fictions").

[7] *See* Krystal Hu, *ChatGPT Sets Record for Fastest-Growing User Base – Analyst Note*, REUTERS (Feb. 2, 2023), https://perma.cc/5AWM-V9FL.

[8] This Essay uses ChatGPT as a shorthand for various AI-enabled chatbots, such as Microsoft Bing, Google Gemini (Bard), and Claude.

[9] *See* Alec Radford, Karthik Narasimhan, Tim Salimans & Ilya Sutskever, Improving Language Understanding by Generative Pre-Training 3 (2018) (unpublished manuscript), https://perma.cc/8F8Q-YM7Y. For a simpler explanation of next-token prediction, see, Alonso Silva Allende, *Next Token Prediction with GPT*, HUGGING FACE (Oct. 20, 2023), https://perma.cc/8VKN-5S7V.

[10] *Id.*

[11] For a detailed explanation of how generative AI and transformer architecture work, see Adam Zewe, *Explained: Generative AI*, MIT NEWS (Nov. 9, 2023), https://perma.cc/5HQ5-H9MY.

[12] *Id.*

[13] *See* Yoshua Bengio et al., Managing AI Risks in an Era of Rapid Progress 2 (Nov. 23, 2023) (unpublished manuscript) (on file with arXiv), https://perma.cc/RK7W-87JS.

This Essay finds that the workings of ChatGPT and the marketplace of ideas theory exhibit at least four common features: arena, means, objectives, and flaws. These shared attributes make ChatGPT the closest realization of the marketplace of ideas theory in history. This comparison also aids us in determining whether truth-seeking should be set as a goal of AI content governance. Part I expounds these four similarities. Part II argues against truth-seeking (or zero-risk) policies and proposes internal and external policy suggestions concerning chatbot output. The contribution of this Essay is two-fold: first, that the internal workings of ChatGPT verify the century-long theoretical discussions of the promise and perils of the marketplace of ideas; second, that various suggestions regarding the reframing of the marketplace theory are also applicable to the development and fine-tuning of LLMs.

## II. COMPARISON

### A. Arena: Information Marketplace

Legal scholars conceive of an ideal information marketplace as a marketplace that encompasses an immense quantity (the volume requirement) of diverse information (the variety requirement). First, the marketplace of ideas requires a sufficient volume of information. Justice Louis Brandeis was the first to justify the constitutional protection of free speech in *Whitney v. California* with the "more speech" principle—"If there be time to expose through discussion the falsehood and fallacies, . . . the remedy to be applied is *more speech*, not enforced silence. Only an emergency can justify repression."[16] The underlying essence of this principle is that a sufficient quantity of information in the marketplace is indispensable for effective competition and the victory of truth over falsehoods.[17] In the century since *Whitney v. California*, the "more speech" principle has gained much influence in legal scholarship and free speech doctrine.[18]

---

[14] *See generally* OpenAI, GPT-4 Technical Report 47-60 (Mar. 27, 2023) (unpublished manuscript) (on file with arXiv), https://perma.cc/XF36-MMY4 [hereinafter OpenAI's Report].
[15] *See generally* Urs Gasser, Navigating AI Governance as a Normative Field: Norms, Patterns, and Dynamics (unpublished manuscript) (on file with author); *see also* Orly Lobel, *The AI Regulatory Pyramid: A Taxonomy & Analysis of the Emerging Toolbox in the Global Race for the Regulation and Governance of Artificial Intelligence*, (San Diego Legal Stud. Paper, Paper No. 24-008), https://perma.cc/7VKN-87WZ.

Second, the marketplace of ideas requires a variety of information.[19] Diverse viewpoints not only foster rigorous competition but also cultivate a dynamic environment where individuals are exposed to conflicting perspectives.[20] This exposure, in turn, enhances the likelihood of individuals making more reasoned decisions, [21] eventually contributing to the overarching goal of the marketplace of ideas—the pursuit of truth. In contrast, homogenous information, even in very large volumes, can by no means catalyze effective competition within the marketplace. Historically, diverse marketplaces included town squares, newspapers, and pamphleting.[22] Today, these arenas have been digitized, facilitating the broader engagement of

---

[16] 274 U.S. 357, 377 (1927) (Brandeis, J., concurring) (emphasis added).

[17] *See* Robert D. Richards & Clay Calvert, *Counterspeech 2000: A New Look at the Old Remedy for "Bad" Speech*, 2000 B.Y.U. L. REV. 553, 553-554 (2000) ("Rather than censor allegedly harmful speech and thereby risk violating the First Amendment's protection of expression, or file a lawsuit that threatens to punish speech perceived as harmful, the preferred remedy is to add more speech to the metaphorical marketplace of ideas."); Brian C. Murchison, *Speech and the Truth-Seeking Value*, 39 COLUM. J.L. & ARTS 55, 116 (2015) ("With the First Amendment, speech enjoys a generous range of freedom because such breadth enhances the chances that accurate (or at least provisionally accurate) understandings will emerge in the aggregate of public discourse.").

[18] Citizens United v. FCC, 558 U.S. 310, 361 (2010) ("The remedies enacted by law, however, must comply with the First Amendment; and it is our law and our tradition that more speech, not less, is the governing rule.").

[19] *See generally* Associated Press v. United States, 326 U.S. 1, 20 (1945) ("[T]he widest possible dissemination of information from *diverse and antagonistic sources* is essential to the welfare of the public, [and] that a free press is a condition of a free society." (emphasis added)).

[20] *See, e.g.*, United States v. Alvarez, 567 U.S. 709, 727 (2012) ("The remedy for speech that is false is speech that is true. . . . The response to the unreasoned is the rational; to the uninformed, the enlightened; to the straight-out lie, the simple truth.").

[21] *See* KENNETH CUKIER, VIKTOR MAYER-SCHÖNBERGER & FRANCIS DE VÉRICOURT, FRAMERS: HUMAN ADVANTAGE IN AN AGE OF TECHNOLOGY AND TURMOIL 177 (2021).

[22] *See* Garrett Morrow & John P. Wihbey, *Marketplace of Ideas 3.0? A Framework for the Era of Algorithms*, 29 RICH. J.L. & TECH. 52, 54 (2022).

Internet users in the information marketplace and the decentralized creation of diverse information.[23]

The performance of LLMs also relies heavily on voluminous and various data—"[t]he more data, the better the program."[24] These data indeed comprise a marketplace quite akin to the theoretical marketplace of ideas. Although GPT-3.5 (i.e., ChatGPT) and GPT-4 (i.e., ChatGPT Plus) have never disclosed the amount of training data they use, the training set details of their predecessor, GPT-3, have been published. GPT-3's training set consists of 45 terabytes of text data, including 410 billion tokens (where one token approximates three-quarters of a word) from Common Crawl (filtered), 19 billion tokens from WebText2, 12 billion tokens from Books1, 55 billion tokens from Books2, and 3 billion tokens from Wikipedia.[25] Roughly speaking, the GPT-3 dataset is equivalent to over 90 million novels of 100,000 words each. Table 1 below reflects the sources of GPT-3's training set.[26]

---

[23] *See* Packingham v. North Carolina, 582 U.S. 98, 107, 137 (2017) ("These websites can provide perhaps the most powerful mechanisms available to a private citizen to make his or her voice heard. They allow a person with an Internet connection to become a town crier with a voice that resonates farther than it could from any soapbox." (internal quotation marks and citation omitted)).

[24] *See, e.g.*, Sara Brown, *Machine Learning Explained*, MIT SLOAN SCH. MGMT. (Apr. 21, 2021), https://perma.cc/FB5A-6H2F ("Machine learning starts with data — numbers, photos, or text, like bank transactions, pictures of people or even bakery items, repair records, time series data from sensors, or sales reports. The data is gathered and prepared to be used as training data, or the information the machine learning model will be trained on. The more data, the better the program.").

[25] *See* Tom B. Brown, Language Models are Few-Shot Learners 8-9 (July 22, 2020) (unpublished manuscript) (on file with arXiv), https://perma.cc/9TR5-BL5D.

[26] *Id.*

Table 1: GPT-3 Training Data, in Billions of Tokens[27]

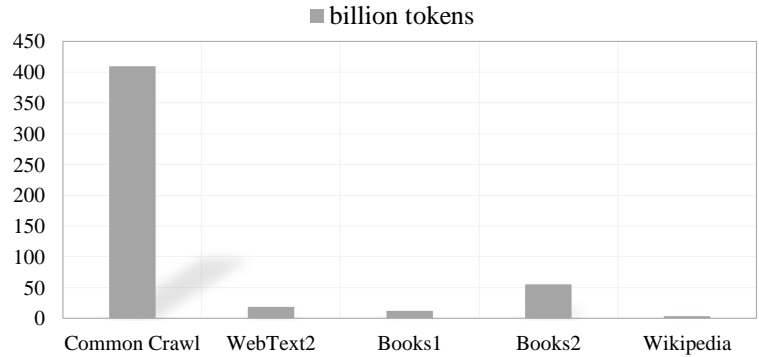

Given the iterative improvements and scale of models in the GPT series, it can be reasonably inferred that GPT-4 was trained on a larger dataset than GPT-3 to enhance its understanding and generative capabilities. According to SemiAnalysis researchers Dylan Patel and Gerald Wong, GPT-4 was trained with approximately 13 trillion tokens.[28] These tokens include text-based and code-based data, as well as data from ScaleAI and internal sources.[29] Other sources may include Twitter, Reddit, textbooks, and newspapers.[30]

### B. Means: Inter-Informational Competition

The efficacy of the marketplace of ideas theory hinges on competition between information for information consumers' attention and acceptance. Here, the competition is between truth and error.[31] Information consumers'

---

[27] *Id.*

[28] *See* Dylan Patel & Gerald Wong, *GPT-4 Architecture, Infrastructure, Training Dataset, Costs, Vision, MoE*, SEMIANALYSIS (July 10, 2023), https://perma.cc/DS5M-8TA9.

[29] *See* Yam Peleg (@Yampeleg), TWITTER (July 11, 2023), https://perma.cc/9CA6-394C.

[30] *Id.*; *see also* Alexandra Bruell, *New York Times Sues Microsoft and OpenAI, Alleging Copyright Infringement*, WALL ST. J. (Dec. 27, 2023), https://perma.cc/5QL3-4YWT.

[31] *See, e.g.*, JOHN MILTON, AREOPAGITICA 51-52 (John W. Hales ed. 1894) ("Let [truth] and falsehood grapple; who ever knew truth put to the worse, in a free and open encounter?"); FREDERICK SCHAUER, FREE SPEECH: A PHILOSOPHICAL ENQUIRY 16 (1982) ("Just as Adam Smith's 'invisible hand' will ensure that the best products emerge from free competition, so too will an invisible hand ensure that the best ideas emerge when all opinions are permitted

judgment thus determines the outcome of such competition.[32] It is information consumers, not speakers, who have the final say on the acceptance of an idea.[33] Subject to certain conditions,[34] the assumption of the marketplace of ideas theory is that truth can consistently triumph by securing *the highest market share*.[35] In other words, the victory of truth over falsehood owes to the truth always attaining the acceptance of the *majority of consumers*. The realization mechanism can be succinctly distilled as follows:

> [T]hat all opinions are to be expressed; everyone comes to the market with his or her ideas, and through discussion everyone exchanges ideas with one another. The ideas or opinions compete with one another, and we have the opportunity to test all of them, weighing one against the other. As rational consumers of ideas, we choose the "best" among them. In the same way that "bad" products naturally get pushed out of the market because of the lack of demand for them and "good" products thrive because they satisfy a demand, so also "good" ideas prevail in the marketplace and "bad" ones are weeded out in due course.[36]

---

freely to compete."); Thomas Jefferson, *A Bill for Establishing Religious Freedom*, 132 DAEDALUS 49, 50 (2003) ("[Truth] is the proper and sufficient antagonist to error, and has nothing to fear from the conflict unless by human interposition disarmed of her natural weapons, free argument and debate; errors ceasing to be dangerous when it is permitted freely to contradict them."); *see also Citizens United*, 558 U.S. at 355 ("Factions should be checked by permitting them all to speak . . . and by entrusting the people to judge what is true and what is false.") (citing THE FEDERALIST NO. 5, at 130 (James Madison) (B. Wright ed., 1961)).

[32] *See* Coase, *supra* note 5, at 27 ("[O]nly if an idea is subject to competition in the marketplace can it be discovered (*through acceptance or rejection*) whether it is false or not." (emphasis added)).

[33] *See* Parsons, *supra* note 5, at 2196-97 ("The listener—not the speaker—decides what content is worthy of attention in a competitive ideational market.").

[34] *See* discussion *infra* Part I.D.

[35] *See* Jill Gordon, *John Stuart Mill and the "Marketplace of Ideas,"* 23 SOC. THEORY & PRAC. 235, 236 (1997).

[36] *Id*.

If we accept this definition and mechanism,[37] truth can assert itself and attain public recognition only if the market share of an idea is calculable or otherwise apparent to consumers; if not, the truth would remain permanently obscured.

The workings of ChatGPT are also statistically based and have actualized the mechanism of the marketplace of ideas. The evolution and capabilities of LLMs like the GPT series from OpenAI can be traced back to the pioneering work on statistical methods for continuous speech recognition by F. Jelinek in 1976.[38] Despite the differences in their specific applications, both LLMs and Jelinek's methods fundamentally leverage statistical learning to process and generate language. Specifically, LLMs are designed to grasp the statistical nuances of language by training on extensive text corpora.[39] This training involves the estimation of the likelihood of various word sequences and their orderings.[40] In text generation, LLMs predict the next word by calculating its conditional probability given the preceding text context. This assessment leads to the selection of the subsequent word either deterministically, by picking the most probable word, or probabilistically, to introduce variety in the generated content.[41]

This means that the frequency of a specific statement within the training data plays a significant role in LLM training and output generation. Patterns and phrases that occur with higher frequency are accorded greater likelihood in similar contexts, guiding the model's understanding and generation of language. This also accounts for why some real-world biases and misinformation are replicated in the output generated by LLMs.[42] In this sense, the inter-informational competition for market share is expressed through the lens of data frequency and the output most accepted by LLMs can be viewed as the so-called "truth" within the marketplace.

---

[37] Here, a paradox lies in the conflicts between the market mechanism and normatively good statements—"[f]ree market theory may assume that it is good for people to formulate and act on their own preferences, but that is not the same as maintaining that those preferences are good ones." Thomas W. Joo, *The Worst Test of Truth: The Marketplace of Ideas as Faulty Metaphor*, 89 TUL. L. REV. 383, 408-09 (2014). For more criticisms, see discussion *infra* Part I.D and Part II.

[38] *See* F. Jelinek, *Continuous Speech Recognition by Statistical Methods*, 64 PROC. IEEE 532 (1976).

[39] *See* Radford et al., *supra* note 9, at 3.

[40] *Id.*

[41] *Id.* at 8.

[42] *See infra* notes 82-86 and accompanying text.

### C.     Objectives: "Truth"-Seeking & Free Speech

The marketplace metaphor is employed for two purposes: to seek truth in the marketplace and to safeguard free speech against government interference. The first objective holds more theoretical than practical significance. Since *Abram v. United States*, courts and scholars have constantly invoked truth-seeking as the ultimate goal of the marketplace of ideas.[43] However, the term "truth" has not been clearly defined in this context. Interpretation of this term is subject to two divergent approaches. The first approach is to view "truth" as a statement that is factually accurate,[44] while the second approach is to place more emphasis on its normative dimension, construing truth as a normatively good idea or opinion.[45]

The second version appears closer to the spirit of the marketplace metaphor, since it is more aligned with Holmes's firm claim that he "do[es]n't believe or know anything about absolute truth."[46] More importantly, only by interpreting "truth" to encompass not only factual statements but also normative opinions can the marketplace metaphor robustly fulfill its role of protecting free speech. Here, a legally significant dichotomy between fact and opinion comes to the fore. In the legal sense, facts and opinions are entitled to different levels of protection. Opinions enjoy more free-speech protections than facts because, opinions, unlike pure facts, are inherently unfalsifiable. [47] Hence, the protection of free speech derived from the marketplace metaphor pertains more to the protection of normative opinions than descriptive facts. This explanation also accounts for why some judges and researchers replace "truth" with terms like "best ideas" or "right conclusions" when quoting the marketplace theory in their arguments.[48]

---

[43] *See, e.g.*, Associated Press v. U.S., 326 U.S. 1, 28 (Frankfurter, J., concurring) ("[R]ight conclusions are more likely to be gathered out of a multitude of tongues, than through any kind of authoritative selection."); Time, Inc. v. Hill, 385 U.S. 374, 406 (1967) (Harlan, J., concurring) ("'The marketplace of ideas' where it functions still remains the best testing ground for truth."); Consol. Edison Co. of New York v. Pub. Serv. Comm'n of New York, 447 U.S. 530, 538 (1980) ("To allow a government the choice of permissible subjects for public debate would be to allow that government control over the search for political truth."); FCC v. League of Women Voters of California, 468 U.S. 364, 377, n.8 (1984) ("[I]t is the purpose of the First Amendment to preserve an uninhibited marketplace of ideas in which truth will ultimately prevail." (citation omitted)); McCullen v. Coakley, 573 U.S. 464, 476 (2014) (reiterating the same notion); Nat'l Inst. of Fam. & Life Advocs. v. Becerra, 858 U.S. 755, 771 (2018) (same); *see also* SCHAUER, *supra* note 31, at 16 (noting that "[t]he numerous characterizations differ from one another . . . but [t]hey all share a belief that freedom of speech is not an end but a means, a means of identifying and accepting truth").

[44] *See* Goldman & Cox, *supra* note 5, at 5 ("What is true or false depends on the way the world actually is, not simply on people's opinions or how they arrive at those opinions.").

[45] *See* Joo, *supra* note 37, at 408.

[46] Blasi, *supra* note 5, at 14 (citing Holmes's words to show that he did not believe in "absolute truth.").

Compared to the first goal, adopting the marketplace metaphor to protect free speech against external interference, especially from governmental regulation,[49] is more practical and implementable. Courts have used the marketplace metaphor extensively in various First Amendment cases to support unfettered information competition and restrict the reach of government in the information marketplace.[50] This stance stems primarily from the distrust of the government—allowing excessive governmental intervention in the informational competition would be equivalent to allowing the government to vote on behalf of the people, which would significantly undermine the foundations of democratic discourse.[51]

---

[47] *See* Gertz v. Robert Welch, Inc., 418 U.S. 323, 340 (1974) ("But there is no constitutional value in false statements of fact. Neither the intentional lie nor the careless error materially advances society's interest in 'uninhibited, robust, and wide-open' debate on public issues." (citation omitted)). *But see* Gettner v. Fitzgerald, 297 Ga. App. 258, 261 (2009) ("[A] statement that reflects an opinion or subjective assessment, as to which reasonable minds could differ, cannot be proved false."); Info. Sys. & Networks Corp. v. City of Atlanta, 281 F.3d 1220, 1228 (11th Cir. 2002) ("Because Commissioner McCall's statement was an opinion—and thus subjective by definition—it is not capable of being proved false.").

[48] *See* Associated Press, 326 U.S., 28 ("[R]ight conclusions are more likely to be gathered out of a multitude of tongues, than through any kind of authoritative selection."); *see also supra* note 36 and accompanying text.

[49] It is debatable whether the purview of the First Amendment should be extended to private sectors that possess great editorial power to intervene in the information market. *See, e.g.*, Kate Klonick, *The New Governors: The People, Rules, and Processes Governing Online Speech*, 131 HARV. L. REV. 1598, 1658-59 (2017) (discussing whether online content platforms are state actors and thus obliged to perform the First Amendment duties).

[50] *See* Smolla, *supra* note 4.

[51] *See* Ingber, *supra* note 5, at 8-12.

However, not all forms of governmental intervention are unnecessary and detrimental to the health of the information market. In fact, the uneven protections afforded to fact and opinion indicate the necessity for the government to rectify basic factual falsehoods in the marketplace. Such interventions are primarily executed through education and endorsement, which are legally acceptable under the First Amendment.[52] Such interventions play a more pivotal role in the construction of social meaning than the regulation of speech because they form the underlying norms of the whole society and establish the rules for the general public to follow.[53]

The development of ChatGPT reflects similar objectives—to generate satisfactory responses (*i.e.*, "truth"-seeking) and to limit excessive content-based intervention (*i.e.*, to protect free speech). Much like the marketplace theory, where "truth"-seeking is not entirely free from governmental involvement, the process of the "truth"-seeking by ChatGPT involves some human interventions. To assist ChatGPT in generating truthful responses, AI experts have employed various technological methods.[54]

First, human actors "educate" and "discipline" ChatGPT. For example, AI experts have introduced a super alignment model, which was trained by a combination of manual annotations and other models' outputs, as a guide to nudge ChatGPT's behavior towards more desirable outcomes.[55] The alignment model is employed to enhance the LLM's understanding and adherence to ethical guidelines and societal norms, reducing the likelihood of generating inappropriate outputs.[56] In addition, human actors also "reward" and "punish" ChatGPT. The reinforcement learning from human feedback (RLHF) approach involves collecting user feedback in the form of precise ratings on ChatGPT's outputs and using these ratings to train a reward model.[57] The reward model evaluates the quality of ChatGPT's responses based on user satisfaction, allowing for the iterative refinement of response quality through a process called policy shaping.[58]

---

[52] *See* Frederick Schauer, *Facts and the First Amendment*, 57 UCLA L. REV. 897, 917-18 (2010).
[53] *See* Lawrence Lessig, *The Regulation of Social Meaning*, 62 U. CHI. L. REV. 943, 973-76 (1995).
[54] *See generally* Luke Munn, Liam Magee & Vanicka Arora, *Truth Machines: Synthesizing Veracity in AI Language Models*, AI & SOC. 1, pt. 3 (2023).
[55] *See* Jan Leike & Ilya Sutskever, *Introducing Superalignment*, OPENAI (July 5, 2023), https://perma.cc/47B7-M4FQ; Collin Burns et al., *Weak-to-Strong Generalization*, OPENAI (Dec. 14, 2023), https://perma.cc/2QAK-NUDS.
[56] *Id*.

Furthermore, human actors also monitor and remove unacceptable content. Rigorous safety and health checks have been in place to proactively filter out prompts or responses that contain potentially harmful content.[59] This is achieved through the development of sophisticated content filtering algorithms that can identify and block a wide range of unhealthy content in real-time, based on predefined criteria related to offensive language, privacy breaches, and misleading information.[60] The use of red teaming language model technology also plays a crucial role in maintaining the integrity of ChatGPT's outputs, which involves continuously monitoring the model's performance to identify and rectify undesirable behaviors.[61] Strategies include listing and blocking responses that contain problematic phrases, identifying and removing offensive content from the training datasets to prevent future occurrences, and employing examples of expected behavior for specific inputs to guide the model's learning process.[62]

However, implementing excessive content-based intervention measures on ChatGPT is just as harmful as allowing unfettered governmental interference in the information marketplace. For example, human feedback on ChatGPT's performance may inherently carry subjective biases, as human evaluators provide evaluations based on their differing perspectives and experiences.[63] Moreover, placing too much emphasis on safety checks could constrain the model's creative capabilities, thus stifling potential innovative developments and possibly unduly restricting the free flow of information.[64] In an environment of excessive intervention, mislabeling content as "harmful" can create false positives and is biased by its very nature.[65] Reverse biases may also arise, where efforts to avoid a previously exposed specific bias (such as racial discrimination) through over-adjustment result in unfair prejudice against other groups (for example, white people).[66] This situation could stem from an overemphasis on political correctness in measures to prevent bias, rather than on balanced and fair information processing.

---

[57] *See, e.g.*, Ryan Lowe & Jan Leike, *Aligning Language Models to Follow Instructions*, OPENAI (Jan. 27, 2022), https://perma.cc/BDQ8-DBD4.

[58] *Id.*

[59] *See generally* Ethan Perez et al., Red Teaming Language Models with Language Models (Feb. 7, 2022) (unpublished manuscript) (on file with arXiv), https://perma.cc/H8C5-9VH7.

[60] *Id.*

[61] *Id.*

[62] *Id.*

[63] *See* Timo Kaufmann et. al., A Survey of Reinforcement Learning from Human Feedback 22 (Dec. 22, 2023) (unpublished manuscript) (on file with arXiv), https://perma.cc/2NAE-MG96.

### D.     *Flaws: Over-Idealistic Assumptions*

The marketplace metaphor has been criticized for decades. The most common challenge centers around the attainability of the ideal that "truth will ultimately prevail." Legal theorists posit that the marketplace of ideas is overly idealistic, arguing that "[it] is as flawed as the economic market."[67] Many scholarly works have argued in favor of this contention. This Essay does not intend to provide an exhaustive list of their arguments.[68] Instead, I select three essential and interrelated flaws to compare with the those of ChatGPT.

First, at the information marketplace level, the marketplace theory assumes that the more information available, the greater the opportunity for truth to surface and prevail, and the better the unregulated marketplace. However, this assumption is highly questionable as the correlation between "more information" and "better results" is far from guaranteed.[69] Researchers have found that recent technological advances greatly favor the creation and dissemination of fake news,[70] and, as a result, today's cyberspace is besieged by disinformation, much of which is maliciously fabricated for political purposes.[71] Even worse, some conspiracy theories gain widespread acceptance and consistently prevail in the competition with the truth,[72] contradicting the anticipated outcomes of an unregulated marketplace, as described in the original marketplace theory.

---

[64] *See, e.g.*, Neel Guha et al., *AI Regulation Has Its Own Alignment Problem: The Technical and Institutional Feasibility of Disclosure, Registration, Licensing, and Auditing*, 92 GEO. WASH. L. REV. (forthcoming 2024) (manuscript at 6), https://perma.cc/Q4HM-GKEH.

[65] *See* OpenAI's Report, *supra* note 14, at 47-48 (noting that "refusals and other mitigations can also exacerbate bias in some contexts, or can contribute to a false sense of assurance").

[66] *See* Thomas Barrabi, *'Absurdly Woke': Google's AI Chatbot Spits Out 'Diverse' Images of Founding Fathers, Popes, Vikings*, N.Y. POST (Feb. 21, 2024), https://perma.cc/XEM9-7MM5; *see also* URS GASSER & VIKTOR MAYER-SCHÖNBERGER, GUARDRAILS: GUIDING HUMAN DECISIONS IN THE AGE OF AI 72-73 (2024) (arguing that "[e]ven sophisticated measures to eliminate one kind of bias can solidify another bias, not just because these measures are insufficient, but because social realities are messy").

[67] Ingber, *supra* note 5, at 16-17.

[68] For a comprehensive review of criticism of the marketplace metaphor, see, for example, PHILIP M. NAPOLI, SOCIAL MEDIA AND THE PUBLIC INTEREST: MEDIA REGULATION IN THE DISINFORMATION AGE 80-106 (2019).

Second, at the societal level, the marketplace theory assumes that different pieces of information are competing on an equal footing, but that is not always the case.[73] In fact, current social structure and information market conditions asymmetrically favor those viewpoints that represent the incumbent, but not necessarily normative good, values while marginalizing those niche perspectives.[74] Ingber astutely identified this flaw and argued that:

> Due to developed legal doctrine and the inevitable effects of socialization processes, mass communication technology, and unequal allocations of resources, ideas that support an entrenched power structure or ideology are most likely to gain acceptance within our current market. Conversely, those ideas that threaten such structures or ideologies are largely ignored in the marketplace.[75]

---

[69] *See generally* Philip M. Napoli, *What If More Speech Is No Longer the Solution? First Amendment Theory Meets Fake News and the Filter Bubble*, 70 FED. COMM. L.J. 55 (2018); *see also* Bambauer, *supra* note 5, at 696-98 (arguing for "more ≠ better").

[70] *See* Napoli, *supra* note 69, at 68-87.

[71] *See* Nunziato, *supra* note 5, at 1527-31.

[72] *See* Schauer, *supra* note 52, at 898.

[73] *See* Ingber, *supra* note 5, at 17-31.

[74] *See id.*

[75] *Id*. at 17.

This cogently accounts for why some conservative ideas that contain biases and stereotypes may gain widespread popularity and dominate the information marketplace for years.

Third, at the consumer level, the marketplace theory assumes that information consumers are rational enough and capable of distinguishing the normative good and bad, truth and falsity. Unfortunately, this appears over-optimistic.[76] Researchers have found that people exhibit very limited rationality when evaluating the substance of information; instead, their judgment is heavily subject to their own political preferences and cultural tastes.[77] Especially in the digital age, technology platforms adopt algorithms to recommend highly targeted and personalized information aligned with users' established preferences.[78] As a result, information consumers are increasingly siloed in their echo chambers and become polarized more easily.[79] Empirical research has also found that even a sensible consumer may have cognitive biases when accessing and evaluating information.[80]

The marketplace theory is theoretically grounded in a virtuous circle assumption—a sufficient variety of information competes equally for the acceptance of consumers who are rational enough to vote for truth and ensure its dominance in the information marketplace, which is presumed to cultivate a healthy information ecosystem and further enlightens future market consumers. However, the overlooked flaws create a vicious circle in the marketplace. As information consumers are inherently biased and irrational, inter-informational competition does not always work effectively to cope with flooding problematic information in the digital space. This will expose future consumers to more problematic information and render them less capable of discerning normative truth.

---

[76] *See, e.g.*, Lyrissa Barnett Lidsky, *Nobody's Fools: The Rational Audience as First Amendment Ideal*, 2010 U. ILL. L. REV. 799, 804 (2010) (indicating that "the marketplace of ideas is flawed because humans are flawed: they are not rational information processors, and more information often leads to worse decisions instead of better ones").

[77] *See* Ingber, *supra* note 5, 31-36; Parsons, *supra* note 5, at 2171-74.

[78] *See* Napoli, *supra* note 69, at 77.

[79] *See* Cass R. Sunstein, *The Law of Group Polarization*, 10 J. POL. PHIL. 175, 185 (2002).

[80] *See* Bambauer, *supra* note 5, at 673-96 (presenting empirical evidence to demonstrate the cognitive biases that people have in accessing and processing information).

Such paradoxes also exist in the development of ChatGPT. First, the datasets that the LLMs were trained on involve a large amount of problematic information. In data science, it has long been known that more data is not necessarily better than higher-quality data.[81] This challenge is paralleled in the training of LLMs, where researchers have found that "[l]arger models are less truthful" supported by empirical evidence.[82] Larger language models run the risk of containing more low-quality data. As previously discussed, many texts widely circulated on the Internet and accessed by LLMs contain factual errors.[83] And, some facts considered accurate before the data cutoff date may later be proven wrong. For example, most of the data for GPT-4 was cut off in April 2023,[84] indicating that events or changes in facts that occurred after that date are not reflected in the model.[85] These problematic inputs can significantly influence LLMs' predictions based on data frequency competition and trigger imitative falsehoods.[86] Hence, increasing the model's size is not necessarily helpful in resolving the imitative weaknesses, as the problem will persist if the training data itself is flawed, irrespective of the model's scale.

Second, many problematic data inputs dominate the datasets and render the competition process of next-token prediction inherently unfair. As mentioned, GPT-4 learns from a vast dataset collected from the Internet, books, articles, and other texts, which inherently carry biases present in human-generated content. If the training data overrepresents or underrepresents certain viewpoints, stereotypes, or demographic statistics, the model may learn and replicate these biases in its predictions.[87] Although GPT-4 is designed to understand and generate human-like text, it does not truly comprehend the context or the ethical implications of its outputs. Its predictions are largely based on the statistical patterns in its training data—"[t]he more frequently a claim appears in the dataset, the higher likelihood it will be repeated as an answer."[88] If these patterns contain biases and misleading content, the model's outputs might also reflect them, namely "bias in, bias out."

---

[81] *See* Jean Boivin & Serena Ng, *Are More Data Always Better for Factor Analysis?* 132 J. ECONOMETRICS 169, 189 (2006). Here, data quality evaluation entails a multidimensional process, including believability, accuracy, objectivity, reputation, value-added, relevancy, timeliness, completeness, appropriate amount, interpretability, ease of understanding, representational consistency, concise representation, accessibility, and access security. *See, e.g.*, Richard Y. Wang & Diane M. Strong, *Beyond Accuracy: What Data Quality Means to Data Consumers*, 12 J. MGMT. INFO. SYS. 5, 14 (1996).

[82] *See* Stephanie Lin, Jacob Hilton & Owain Evans, TruthfulQA: Measuring How Models Mimic Human Falsehoods 2-3, 6-7 (May 8, 2022) (unpublished manuscript) (on file with arXiv), https://perma.cc/9RZF-LCV7.

[83] *See supra* notes 70-71 and accompanying text.

[84] *See* OpenAI, *OpenAI DevDay: Opening Keynote*, YOUTUBE (Nov. 6, 2023), https://perma.cc/ZF9S-79QG.

[85] *See* OpenAI's Report, *supra* note 14, at 58.

[86] *See generally* Lin, Hilton & Evans, *supra* note 82.

Third, users or flaggers of ChatGPT are not rational and cannot furnish perfectly objective ratings for ChatGPT's output. Just as the information consumers engage in the competition of the information marketplace by accepting, creating, and disseminating an idea, users or flaggers of ChatGPT do the exact same thing by evaluating ChatGPT's responses. However, the RLHF and super alignment model, as essential tools to enhance the output quality, are unavoidably influenced by the personal bias of the participants.[89] Feedback is highly contingent on an individual's identity, religious beliefs, educational backgrounds, and personal history.[90] Despite efforts to diversify feedback, completely eliminating subjective bias remains a challenge. Especially, if the super alignment technology is primarily driven by specific companies or sectors, it could lead to these entities exerting too much influence over the values of AI chatbots, thereby affecting the collective values of humanity as a whole.[91]

---

[87] *See* Laura Weidinger et al., Ethical and Social Risks of Harm from Language Models 11 (Dec. 8, 2021) (unpublished manuscript) (on file with arXiv), https://perma.cc/6FQV-B9HF.

[88] Munn, Magee & Arora, *supra* note 54, at 3.

[89] *See* Kaufmann et. al., *supra* note 63, at 22.

[90] *See* Long Ouyang et al., Training Language Models to Follow Instructions with Human Feedback 19 (Mar. 4, 2022) (unpublished manuscript) (on file with arXiv), https://perma.cc/5F87-NV62.

[91] *See id.* at 20; *see also* UC Berkeley Events, *Excavating "Ground Truth" in AI: Epistemologies and Politics in Training Data*, YOUTUBE (Mar. 8, 2022), https://perma.cc/9JQK-4QYQ.

Table 2: Comparing Marketplace of Ideas & Workings of ChatGPT

|  | *Marketplace of Ideas* | *Workings of ChatGPT* |
|---|---|---|
| **Arena** | Marketplace with voluminous and various information | Datasets including voluminous and various data |
| **Means** | Information competes for market share in the marketplace | Data compete for frequency in next-token predictions |
| **Objectives** | To seek truth, especially the normative truth (with some government involvement, *e.g.*, education and endorsement) | To generate satisfactory responses, or "truth" (with some human involvement, *e.g.*, super alignment model, RLHF approach, and monitoring and cleansing) |
| **Objectives** | To protect free speech against excessive intervention from government | To prevent excessive human intervention since it can create subjective inaccuracy, false positives, and reverse biases |
| **Flaws** | "more [information]≠better [outcome]" | "Larger models are less truthful." |
| **Flaws** | "[I]deas that support an entrenched power structure or ideology are most likely to gain acceptance." | "The more frequently a claim appears in the dataset, the higher likelihood it will be repeated as an answer." |
| **Flaws** | "[T]he marketplace of ideas is flawed because humans are flawed: they are not rational information processors . . . ." "[A] growing scientific consensus reveals intuitive beings construing content based on relationships, associations, social identities, and innate biases." | "[RLHF] is attended by all-too-human subjectivity." "Some of the labeling tasks rely on value judgments that may be impacted by the identity of our contractors, their beliefs, cultural backgrounds, and personal history." |

### III. SUGGESTIONS

Despite the potential for imperfect analogies, ChatGPT undeniably stands out as the engine that has come closest to presenting the full picture of the marketplace theory to date. Thus, if we endorse the values of the marketplace of ideas in enhancing the free flow of information, we should also welcome the chatbot engines actualizing these values. At the same time, we should also reflect on our governance models for ChatGPT, just akin to the scrutiny directed at the overly idealized nature of the marketplace theory. Against this backdrop, I argue that those accrued reflections and strategies that academics have designed for reframing marketplace theory are also applicable to the realm of AI chatbot governance. Thus, this Essay generates the following suggestions for AI companies' internal alignments and policy agencies' external governance.

First, expecting and requiring chatbot companies to fully de-risk their systems and produce flawless responses (or "truth") is a fantasy at most. Currently, researchers and engineers are adamant about developing various strategies to mitigate AI content risks by refusing certain prompts[92] and to align AI output with the "truth."[93] Worldwide policymakers also propose heavy-handed rules to mitigate the risks posed by AI output, aspiring to realize the truth-seeking goal in AI output.[94] However, an overarching question faced by both the marketplace metaphor and AI chatbot governance is: although we assume the normative truth exists, [95] is truth-seeking attainable in the idea (not fact)[96] marketplace, and is it appropriate to set it as an ultimate objective? The initial version of marketplace theory offers little guidance here. Holmes tautologically defines "truth" merely as the idea that prevails in the information marketplace without referring to any substantive and ethical meaning of the message conveyed.[97] But, it is far from the truth that truth is tantamount to whatever prevails in the marketplace or whatever is generated by AI models.

---

[92] *See* OpenAI's Report, *supra* note 14, at 44-51.

[93] *See* Munn, Magee & Arora, *supra* note 54, at pt. 3; *see also* Roth, *supra* note 2 ("Elon Musk says he's working on 'TruthGPT,' a ChatGPT alternative that acts as a 'maximum truth-seeking AI.'"); TruthGPT, https://perma.cc/CNK8-E8WV (last visited Apr. 15, 2024) (claiming that "truthgpt's ai chatbot is truth-oriented, meaning it gives responses that are geared towards being unbiased and the closest to the truth as possible").

[94] For example, the Chinese government has demonstrated a pronounced interest in a zero-risk policy and set an impressively long list of prohibited outputs in its Interim Measures. *See* Shengcheng Shi Rengong Zhineng Fuwu Guanli Zanxing Banfa (生成式人工智能服务管理暂行办法) [Interim Measures for Regulating Generative AI Services] (promulgated by Cyberspace Admin., Nat'l Dev. & Reform Comm'n, Ministry Educ., Ministry Sci. & Tech., Ministry Indus. & Info. Tech., Ministry Pub. Sec., Nat'l Radio & Television Admin., effective Aug. 15, 2023), Art. 4, https://perma.cc/H52E-N99R. For an English-translated version, see, for example, Interim Measures for the Management of Generative Artificial Intelligence Services, CHINA L. TRANSLATE (July 13, 2023), https://perma.cc/KR3Z-GW9M.

This Essay argues that it is ill-advised to set "truth"-seeking as the goal of both marketplace theory and chatbot governance. Actually, we never expect a so-called "best commodity" to dominate the physical market; what we value is people having sufficient autonomy to choose commodities according to their own preferences. Why should we harbor such an expectation that "truth will prevail" in the information marketplace? The

---

Other jurisdictions, such as the European Union, also exhibited similar impulses in its legislative process. *See* Amendments Adopted by the European Parliament on 14 June 2023 on the Proposal for a Regulation of the European Parliament and of the Council on Laying Down Harmonised Rules on Artificial Intelligence (Artificial Intelligence Act) and Amending Certain Union Legislative Acts, §28b(4)(b) (COM(2021)0206 – C9-0146/2021 – 2021/0106(COD)), https://perma.cc/HB8E-MLZK (requiring generative AI providers to "train, and where applicable, design and develop the foundation model in such a way as to ensure adequate safeguards against the generation of content in breach of Union law in line with the generally-acknowledged state of the art, and without prejudice to fundamental rights, including the freedom of expression"). This provision, however, was not adopted in the final version of the EU AI Act.

[95] At the philosophical level, there are considerable disputes surrounding the existence of normative truth. *See generally* SCOTT SOAMES, UNDERSTANDING TRUTH (1999); *see also* RONALD DORKIN, JUSTICE FOR HEDGEHOGS, at pt.1 (2011). However, there is scant disagreement regarding the formidable challenges inherent in the attainability of normative truth.

[96] For the dichotomy between fact and idea (or opinion), see *supra* notes 47-48 and accompanying text.

[97] Some researchers comment that Holmes's interpretation of "truth" is no more than a tautology. *See* CASS R. SUNSTEIN, DEMOCRACY AND THE PROBLEM OF FREE SPEECH 25 (1993) ("Truth itself is defined by reference to what emerges through 'free trade in ideas.' For Holmes, it seems to have no deeper status."); *see also* Joo, *supra* note 37, at 406, n.161.

acceptance of a statement, much like the acceptance of a commodity, depends on multifaceted factors far beyond the sole metric of accuracy.[98]

Alternatives should be proposed here. Understanding the challenges of seeking absolute "truth" in the marketplace of ideas, some researchers argue that "[t]he value that is to be realized is not in the possible attainment of truth, but rather, in the existential value of the search itself."[99] Professor Joseph Blocher takes a step further and proposes a knowledge-based approach to rectify the epistemological inappropriateness of the truth-seeking ideal.[100] In Blocher's view, "the goal of free speech is not the maximization of truths in the abstract, but rather the development of knowledge."[101] What truly matters is the underlying justifications that support ideas, instead of mere accuracy.[102] This proposal holds significant heuristic value for AI chatbot governance.

An appealing direction for LLMs would be to train them not only to generate direct responses to the user's prompts but also to present the justifications underpinning these responses. In other words, the desirable chatbot output should not converge towards a single, absolute truth, but rather diverge towards various perspectives based on sufficient justifications. This approach advocates for a falsity-tolerant governance strategy since falsity, although not welcomed, may become acceptable if it is justified with sufficient reasons.[103] Therefore, a preferable way to mitigate the inherent content risks is to present diverse, competing ideas in a specific marketplace, instead of broadening the purview of refusal to certain requests.[104]

---

[98] *See* Schauer, *supra* note 52, at 909; Parsons, *supra* note 5, at 2159 ("And where the Court envisions calculating individuals dispassionately comparing and contrasting information in a vacuum, a growing scientific consensus reveals intuitive beings construing content based on relationships, associations, social identities, and innate biases.").

[99] William P. Marshall, *In Defense of the Search for Truth as a First Amendment Justification*, 30 GA. L. REV. 1, 4 (1995).

[100] *See generally* Joseph Blocher, *Free Speech and Justified True Belief*, 133 HARV. L. REV. 439 (2019).

[101] *Id.* at 459.

[102] *Id.* at 464.

[103] *Id.* at 481-82 (arguing that "a false statement is deserving of constitutional coverage when it is based on a sufficient justification").

Remarkably, some AI researchers have initiated the first strides along this promising trajectory very recently. For instance, some researchers introduced a PRISM alignment project to map detailed survey responses of humans from diverse backgrounds to 8,011 live conversations with 21 LLMs.[105] Their findings reveal the significance of including pluralistic voices for the welfare of the general public and open doors for upcoming research in developing a more desirable alignment approach.[106] Similarly, some researchers, realizing that RLHF is wrongfully based on the assumption that LLMs should be aligned to the "average" human preference, propose a roadmap to pluralistic alignment, which is "capable of representing a diverse set of human values and perspectives."[107] They generate and evaluate three different methods, namely Overton, Steerable, and Distributional models, to operationalize the pluralism in the chatbot output.[108] It will be even more impressive, in my view, if LLMs are trained to generate pluralistic opinions based on sufficient and concrete justifications. To create a more inclusive environment, AI policies should also shift from imposing harsher risk clearance or truth-seeking duties to taking a lenient approach to encourage AI systems to represent a wide range of perspectives, values, and cultural contexts.

---

[104] For examples of model refusals aimed to reduce the tendency of language models to produce harmful content, see OpenAI's Report, *supra* note 14, at 49.

[105] Hannach Rose Kirk et al., The PRISM Alignment Project: What Participatory, Representative and Individualised Human Feedback Reveals About the Subjective and Multicultural Alignment of Large Language Models (Apr. 24, 2024) (unpublished manuscript) (on file with arXiv), https://perma.cc/4WE4-HETF.

[106] *Id.*

[107] *See generally* Taylor Sorensen et al., A Roadmap to Pluralistic Alignment (Feb. 7, 2024) (unpublished manuscript) (on file with arXiv), https://perma.cc/9EK5-HG6M (proposing and designing a novel alignment model to serve pluralistic human values).

[108] *Id.* at 2-7.

Figure 3: Three Kinds of Pluralism in Models[109]

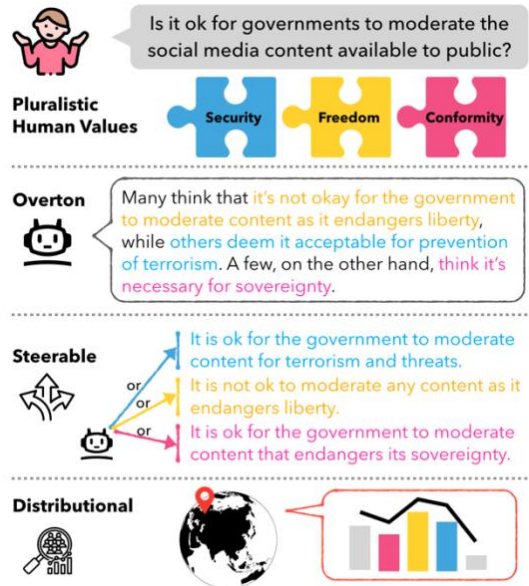

This does not mean that de-risking generative AI systems is not a pressing issue. The burden of de-risking generative AI systems, however, should be appropriately allocated to different social actors. This is essentially because the content risks "posed" by ChatGPT are actually rooted in the whole information market.[110] In this sense, ChatGPT is just reflecting, instead of creating, the problematic information that has already existed in the marketplace. It is therefore unjust to solely burden chatbot companies with de-risking duties. The mitigation of AI content risk necessitates the systematic moderation of the entire information ecosystem. This is a protracted and cross-cutting endeavor, from education and training initiatives focusing on improving digital literacy and AI literacy to governmental policies that are proportionately tailored to enhance inter-informational competition.

---

[109] *Id.* at 1.
[110] *See supra* notes 87-88 and accompanying text.

IV.  CONCLUSION

As an enduring metaphor within legal discourse, the marketplace of ideas theory has wielded considerable influence over the jurisprudential landscape for decades. A century following the inception of this theory, ChatGPT emerged as a revolutionary technological advancement in the twenty-first century, effectively manifesting the marketplace metaphor—it not only substantiates the promises envisaged by generations of legal scholars but also lays bare the perils discerned through sustained academic critique.

The comparison of the marketplace theory and ChatGPT merely marks a starting point. A more meaningful undertaking entails reevaluating and reframing both internal and external AI policies by referring to the accumulated experience, insights, and suggestions researchers have raised to fix the marketplace theory. Here, a pivotal issue is: should truth-seeking be set as the goal of AI content governance? This Essay suggests that, given the unattainability of the goal to seek absolute truth, adopting zero-risk policies proves ill-advised. Instead, a more judicious approach would be to embrace a knowledge-based alternative wherein LLMs are trained to generate competing and divergent viewpoints based on sufficient justifications. This Essay also argues that so-called AI content risks are not created by AI companies but are inherent in the entire information ecosystem. Thus, the burden of managing these risks should be distributed among different social actors, rather than being solely shouldered by chatbot companies.

Last but not least, this Essay focuses on the internal workings of ChatGPT and argues that at this granular level, ChatGPT acts as an engine actualizing the marketplace of ideas. At a more macroscopic scale, the output of ChatGPT competes with other information outlets for quality, relevance, accuracy, and other aspects, constituting a new and broader marketplace.[111] We should also be confident that some risks posed by ChatGPT can be mitigated by inter-informational competition in the marketplace. Only those risks that remain unaddressed by market forces necessitate government intervention.[112] This suggestion aligns with the ethos of the First Amendment

---

[111] *See generally* Jiawei Zhang, *Regulating Chatbot Output via Inter-Informational Competition*, 22 Nw. J. Tech. & Intell. Prop. (forthcoming 2024), https://perma.cc/H4UP-FKCR.
[112] *Id.*

to protect the people's right to interact with technologies and access information.[113]

---

[113] *See generally* Eugene Volokh, Mark A Lemley & Peter Henderson, *Freedom of Speech and AI Output*, J. Free Speech L. 651 (2023); Cass R. Sunstein, *Cass R. Sunstein: "Does Artificial Intelligence Have the Right to Freedom of Speech?"*, Network L. Rev. (Feb. 28, 2024), https://perma.cc/5PLZ-5WGH.